\pdfoutput=1

\documentclass[11pt]{article}

\PassOptionsToPackage{sort}{natbib}

\usepackage{ACL2023}
\usepackage{booktabs}
\usepackage[inline]{enumitem}
\usepackage[skip=0pt]{caption}
\usepackage{acronym}
\usepackage{multirow}
\usepackage{makecell}
\usepackage{graphicx}
\usepackage{times}
\usepackage{xcolor}
\usepackage{subcaption}
\usepackage{wrapfig}
\usepackage{amsmath}
\usepackage{xspace}

\usepackage{times}
\usepackage{latexsym}

\usepackage[T1]{fontenc}

\usepackage[utf8]{inputenc}

\usepackage{microtype}

\usepackage{inconsolata}

\AtBeginDocument{%
  \providecommand\BibTeX{{%
    \normalfont B\kern-0.5em{\scshape i\kern-0.25em b}\kern-0.8em\TeX}}}

\acrodef{DS}{dialogue system}
\acrodef{TDS}{task-oriented dialogue system}
\acrodef{ReDial}{recommendation dialogue}
\acrodef{IR}{information retrieval}
\acrodef{MTurk}{Amazon Mechanical Turk}
\acrodef{CRS}{conversational recommender system}
\acrodef{RS}{recommender system}
\acrodef{CR}{conversation recommendation}
\acrodef{LLM}{large language model}
\acrodef{HIT}{human intelligence task}
\acrodef{IAA}{inter-annotator agreement}

\newcommand{\header}[1]{\vspace{1mm}\noindent\textbf{#1.}}
\newcommand{\Czero}{$C_{0}$\xspace}
\newcommand{\Cthree}{$C_{3}$\xspace}
\newcommand{\Cseven}{$C_{7}$\xspace}

\author{Clemencia Siro \qquad Mohammad Aliannejadi \qquad Maarten de Rijke \\
        University of Amsterdam, Amsterdam, The Netherlands\\  
        \texttt{\{c.n.siro,m.aliannejadi,m.derijke\}@uva.nl}}

\begin{document}

\title{Context Does Matter: Implications for Crowdsourced Evaluation\\ Labels in Task-Oriented Dialogue Systems}

\maketitle

\begin{abstract}
Crowdsourced labels play a crucial role in evaluating \acp{TDS}. Obtaining high-quality and consistent ground-truth labels from annotators presents challenges. When evaluating a \ac{TDS}, annotators must fully comprehend the dialogue before providing judgments. Previous studies suggest using only a portion of the dialogue context in the annotation process. However, the impact of this limitation on label quality remains unexplored. This study investigates the influence of dialogue context on annotation quality, considering the truncated context for relevance and usefulness labeling. 
We further propose to use \acp{LLM} to summarize the dialogue context to provide a rich and short description of the dialogue context and study the impact of doing so on the annotator's performance.
Reducing context leads to more positive ratings. Conversely, providing the entire dialogue context yields higher-quality relevance ratings but introduces ambiguity in usefulness ratings. 
Using the first user utterance as context leads to consistent ratings, akin to those obtained using the entire dialogue, with significantly reduced annotation effort. Our findings show how task design, particularly the availability of dialogue context, affects the quality and consistency of crowdsourced evaluation labels.\footnote{To foster research in this area, we release our data publicly at \url{https://github.com/Clemenciah/Effects-of-Dialogue-Context}}
\end{abstract}

\acresetall

\section{Introduction}

With recent advances in pre-trained language models and \acp{LLM}, \acfp{TDS} have redefined how people seek information, presenting a more natural approach for users to engage with information sources~\citep{budzianowski-hellogpt2,TOD-BERT}. 
As~\acp{TDS} become increasingly integral to information-seeking processes, the question of how to accurately and effectively evaluate their performance becomes critical. 
Due to the poor correlation of automatic metrics with human-generated labels~\citep{Deriu2020SurveyOE}, evaluation of \acp{TDS} has shifted towards relying on user ratings or crowdsourced labels as ground-truth measures~\citep{acute-eval}.\looseness=-1 

Various crowdsourcing techniques have been employed to collect ground-truth labels, such as sequential labeling~\citep{simulating-usat}, where the annotators go through each utterance and annotate them one by one. This approach introduces certain risks in the annotation process, such as annotators' fatigue and high cognitive load in extra-long dialogues, requiring them to remember and track the state of the dialogue as they annotate the utterances~\citep{clemencia-sat}.
While following and understanding the dialogue context is crucial and can influence the annotators' ratings,
reading and understanding very long dialogues can lead to degraded performance.

To address this issue, another line of research proposes to randomly sample only a few utterances in each dialogue to be annotated~\citep{clemencia-sat,mehri-eskenazi-2020-usr,Siro-UsatCRS}. While addressing the high cognitive load and fatigue, limiting annotators' understanding of the dialogue poses obvious risks, such as unreliable and biased labels~\cite{clemencia-sat,Schmitt-quality-of-interaction}. 
In particular, the amount of dialogue context can lead to biases. 
For example, annotators who lack rich context may unintentionally lean towards positive or negative ratings, neglecting the broader quality of the response. 
Thus, offering annotators too little context risks misleading judgments, potentially leading to inaccurate or inconsistent labels. Conversely, flooding annotators with excessive information can overwhelm them, which can lead to lower returns in terms of label quality. 

Prior work has investigated factors that affect the quality and consistency of crowdsourced evaluation labels, including annotator characteristics, task design, cognitive load, and evaluation protocols~\citep[see, e.g.,][]{DBLP:journals/ipm/RoiteroMMS21,DBLP:conf/eacl/ParmarMGB23,DBLP:cognitiveload-ds,DBLP:conf/sigir/RoiteroSFSMD20}.
However, no previous work studies the effect of random sampling and the number of sampled utterances on the annotation quality.

In this study, we aim to address this research gap by investigating how different amounts of contextual information impact the quality and consistency of crowdsourced labels for \acp{TDS}, contributing to understanding of the impact of such design choices.
We experiment with crowdsourcing labels for two major evaluation aspects, namely, \emph{relevance} and \emph{usefulness} under different conditions, where we compare the annotation quality under different dialogue context truncation strategies.

Addressing the challenge of insufficient context at the turn level, we propose to use heuristic methods and \acp{LLM} to generate the user's information need and dialogue summary. 
\acp{LLM} can play the role of annotation assistants~\cite{faggioli2023perspectives} by summarizing the dialogue history, facilitating a more efficient and effective understanding of the dialogue context before annotating an utterance. To this aim, we use GPT-4 for dialogue context summarization and compare the performance of annotators' under different conditions, as well as different context sizes.
Through these experiments, we answer two main questions:
\begin{enumerate*}[label=\textbf{(RQ\arabic*)}]
     \item How does varying the amount of dialogue context affect the crowdsourced evaluation of \acp{TDS}?
     \item Can the consistency of crowdsourced labels be improved with automatically generated supplementary context?\looseness=-1
\end{enumerate*} 

Our findings reveal that the availability of previous dialogue context significantly influences annotators' ratings, with a noticeable impact on their quality. 
Without prior context, annotators tend to assign more positive ratings to system responses, possibly due to insufficient evidence for penalization, introducing a positivity bias. 
In contrast, presenting the entire dialogue context yields higher relevance ratings. 
As for usefulness, presenting the entire dialogue context introduces ambiguity and slightly lowers annotator agreement. This highlights the delicate balance in contextual information provided for evaluations. 
The inclusion of automatically generated dialogue context enhances annotator agreement in the no-context~(\Czero) condition while reducing annotation time compared to the full-context~(\Cseven) condition, presenting an ideal balance between annotator effort and performance.

Our findings extend to other task-oriented conversational tasks like conversational search and preference elicitation, both relying on crowdsourced experiments to assess system performance.

\section{Methodology}

We examine how contextual information about a dialogue affects the consistency of crowdsourced judgments regarding \textit{relevance} and \textit{usefulness} of a dialogue response. 
Here, contextual information refers to the information or conversation that precedes a specific response.
We carry out experiments in two phases.  
\textbf{Phase~1} involves varying the \emph{amount} of dialogue context for annotators to answer \textbf{RQ1}.
In \textbf{Phase~2}, we vary the \emph{type} of previous contextual information available to annotators to address \textbf{RQ2}.

\subsection{Experimental data and tasks}

We use the \ac{ReDial} dataset~\citep{li2018conversational}, a conversational movie recommendation dataset, comprising of over 11K dialogues. The dataset is collected using a human-human approach, i.e., one person acts as the movie seeker, while the other is the recommender with the goal of recommending a suitable movie to the seeker, thus making the dataset goal-oriented.
We randomly select system responses from 40 dialogues for the assignment of relevance and usefulness labels. 
These dialogues typically consist of 10 to 11 utterances each, with an average utterance length of 14 words. 
We evaluate the same system responses across all experimental conditions.

The annotation task for the annotators involves two dimensions:
\begin{enumerate*}[label=(\roman*)]
    \item \emph{relevance}: Is the system response relevant to the user's request, considering the context of the dialogue? And
    \item \emph{usefulness}: How useful is the system's response given the user's information need?
\end{enumerate*}
For the \textit{relevance task} we ask annotators to judge how relevant the system's recommendations are to the user's request~\citep{DBLP:journals/sigir/AlonsoRS08}. 
First, the annotator has to judge whether the system response includes a movie recommendation or not; if yes, the annotator assesses whether the movie meets the user's preference; if not, we ask them to note that the utterance does not recommend a movie.
The judgment is on a binary scale for the latter case, where the movie is either relevant (1) or not (0).  
For each experimental condition (see below), annotators only assess the system response with access to the previous context. 
Note that we forego the user's feedback on the evaluated response~(next user utterance) so as to focus on topical relevance of the recommended movie, that is, if the movie meets the user request and preference in terms of the genre, actor, director, etc.
For the \emph{usefulness task} annotators assess a response with or without a movie recommendation with the aim of determining how useful the system's response is to the user~\citep{DBLP:conf/sigir/MaoLZNSZMSL16}. 
The judgment is done on a three-point scale~(i.e., very, somewhat, and not useful). 
Unlike the relevance task, annotators have access to the user's next utterance for the usefulness task; usefulness is personalized to the user, in that even though a movie may be in the same genre, sometimes a user may not like it (e.g., does not like the main actor), thus making the system response relevant but not useful to the user.

\subsection{Automatic generation of diverse dialogue contexts} 
\label{section:additional-context}
\textbf{User information need.}
The user's information need plays a significant role when assessing or improving the quality of the data collected in \acs{IR} systems~\citep{DBLP:conf/sigir/MaoLZNSZMSL16}. 
It refers to \textit{the specific requirement or query made by a user, which guides the system in understanding their preferences and retrieving relevant information to fulfill that need}.
For \acp{TDS}, understanding the user's intent is crucial for annotators participating in the evaluation, as they are not the actual end users. 
This understanding improves the alignment of evaluation labels with the actual user's requirements. 
We define the user's information need as their movie recommendation preference.
Given the consistency of user preferences in the ReDial dataset, where users tend to maintain a single preference throughout a conversation, providing the user's initial information need aids annotators in evaluating the current turn for relevance or usefulness. 

We adopt two approaches to generate the user's information need.
One is to heuristically extract the first user utterance that either requests a movie recommendation or expresses a movie preference, based on phrases such as ``looking for,'' ``recommend me,'' and ``prefer.'' 
These phrases are extracted from the first three user utterances in a dialogue, with the top 10 most common phrases selected.
The second approach relies on \acp{LLM} to generate the user's information need. 
We hypothesize that \acp{LLM} can identify pertinent user utterances in a dialogue and generate the corresponding information need. 
We use GPT-4~\citep{openai-gpt4-report} in a zero-shot setting; with the dialogue context up to the current turn as input, we prompt the model to generate the user's information need.

\header{Generating dialogue summaries}
Dialogue summarization is beneficial for providing a quick context to new participants of a conversation and helping people understand the main ideas or search for key contents after the conversation, which can increase efficiency and productivity~\citep{DBLP:conf/ijcai/FengF022}. 
We use dialogue summaries to provide annotators with quick prior context of a dialogue.
We use GPT-4~\citep{openai-gpt4-report} in a zero-shot setting, as in the case of user information needs, but vary the prompt. 
We instruct GPT-4 to generate a summary that is both concise and informative, constituting less than half the length of the input dialogue. 
Both the generated user information needs and summaries are incorporated in Phase~2 of the crowdsourcing experiments.

Due \acp{LLM}' potential for hallucination~\citep{LLM-survey,DBLP:conf/emnlp/Bouyamourn23}, we evaluate the generated summaries and user information need to ensure factuality and coherence. We elaborate the steps we took in Section~\ref{datacontrol}.

\subsection{Crowdsource experiments}
 
Following~\citep{DBLP:journals/ir/KazaiKM13,DBLP:conf/sigir/RoiteroSFSMD20,DBLP:conf/ecir/Kazai11}, we design \ac{HIT} templates to collect relevance and usefulness labels. 
We deploy the \acp{HIT} in variable conditions to understand how contextual information affects annotators' judgments. 
Our study has two phases: in Phase~1 we vary the \emph{amount} of contextual information; in Phase~2 we vary the \emph{type} of contextual information. 
In each phase and condition, the annotators were paid the same amount as this study is not focused on understanding how incentive influences the quality of crowdsourced labels. 
Like~\citep{DBLP:journals/ir/KazaiKM13}, we refrain from disclosing the research angle to the annotators in both phases; this helps prevent potential biases during the completion of the \ac{HIT}.

\header{Phase~1}\label{phase1} In Phase~1, the focus is on understanding how the \emph{amount} of dialogue context impacts the quality and consistency of relevance and usefulness labels. 
We vary the length of the dialogue context to address (RQ1). 
Thus, we design our experiment with three variations: \Czero, \Cthree, and \Cseven (see Section~\ref{Variations}). The \ac{HIT} consists of a general task description, instructions, examples, and the main task part.
For each variation, we gather labels for two main dimensions (relevance and usefulness) and include an open-ended question to solicit annotators' feedback on the task. 
Each dimension is assessed with 3 annotators in a separate \ac{HIT}, with the same system response evaluated by each. This ensures a consistent evaluation process for both relevance and usefulness. 

\header{Phase~2} In Phase~2, the focus shifts to the \emph{type} of contextual information, to answer (RQ2). We take an approach of machine in the loop for crowdsourcing. 
We restrict our experiments to experimental variation \Czero (defined below), where no previous dialogue context is available to the annotators.
We aim to enhance the quality of crowdsourced labels for \Czero by including additional contextual information alongside the turn being evaluated. 
Our hypothesis is that without prior context, annotators may face challenges in providing accurate and consistent labels. 
By introducing additional context, like the user's information need or a dialogue summary, we expect an increase in the accuracy of evaluations. Through this, we aim to approach a level of performance similar to when annotators have access to the entire dialogue context while minimizing the annotation effort required. We enhance the 40 dialogues from Phase~1 with the user's information need or a dialogue summary, as detailed in Section~\ref{section:additional-context}. Thus, in Phase~2, we have three experimental setups: \Czero -llm, \Czero -heu, and \Czero -sum. 
Table~\ref{tab:variations} in Section~\ref{experimentalconditions} summarizes the setups.

The \ac{HIT} design closely mirrors that of Phase~1. The main task remains unchanged, except for the inclusion of the user's information need or a dialogue summary. Annotators answer the same two questions on relevance and usefulness in separate \ac{HIT}s. While we do not strictly enforce reliance on the additional information provided, annotators are encouraged to use it when they perceive that the current response lacks sufficient information for an informed judgment.

\subsection{Experimental conditions}
\label{Variations}

We focus on two key attributes: the \emph{amount} and \emph{type} of dialogue context. 
For both attributes, we explore three distinct settings, resulting in 6 variations, for both relevance and usefulness; each was applied to the same 40 dialogues:
\begin{itemize}[leftmargin=*,nosep]
    \item \textit{Amount of context}. We explore three truncation strategies: no-context~(\Czero), partial context~(\Cthree), and full context~(\Cseven), designed to encompass scenarios where no previous dialogue context is accessible to the annotator (\Czero), where some previous dialogue context is available but not comprehensively (\Cthree), and when annotators have access to the complete previous dialogue context (\Cseven).
    \item \textit{Type of context}. Using the contexts generated in Section~\ref{section:additional-context}, we experiment with three variations of context type: heuristically generated information need  (\Czero -heu), an \ac{LLM}-generated information need (\Czero -llm), and dialogue summary (\Czero -sum). 
\end{itemize}
Table~\ref{tab:variations} in Section~\ref{experimentalconditions} of the appendix summarizes the experimental conditions.

\subsection{Participants}

We enlisted master workers from the US on \ac{MTurk}~\citep{mturk} to ensure proficient language understanding. Annotators were filtered based on platform qualifications, requiring a minimum accuracy of 97\% across 5000 HITs. To mitigate any learning bias from the task, each annotator was limited to completing 10 HITs per batch and participating in a maximum of 3 experimental conditions. A total of 78 unique annotators took part in Phases 1 and 2 and each worker was paid \$0.4 per HIT, an average of \$14 per hour. Their average age range was 35--44 years. The gender distribution was $46\%$ female and $54\%$ male. 
The majority held a four-year undergraduate degree ($48\%$), followed by two-year and master's degrees ($15\%$ and $14\%$, respectively).

We conduct quality control on the crowdsourced labels to ensure reliability as described in Section~\ref{datacontrol} in the appendix.

\section{Results and Analysis}

We address \textbf{(RQ1)} and \textbf{(RQ2)} by providing an overview of the results and in-depth analysis of our crowdsourcing experiments.  We first describe the key data statistics.

\subsection{Data statistics}

\begin{figure}[!t]
    
    \centering
    \subfloat[Relevance]{\includegraphics[width=0.5\columnwidth]{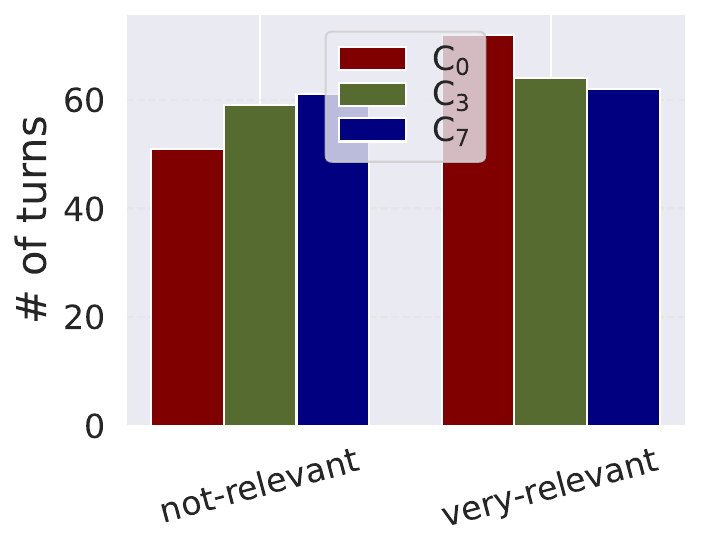}\label{fig:relevant-phase1}}
    \subfloat[Usefulness]{\includegraphics[width=0.5\columnwidth]{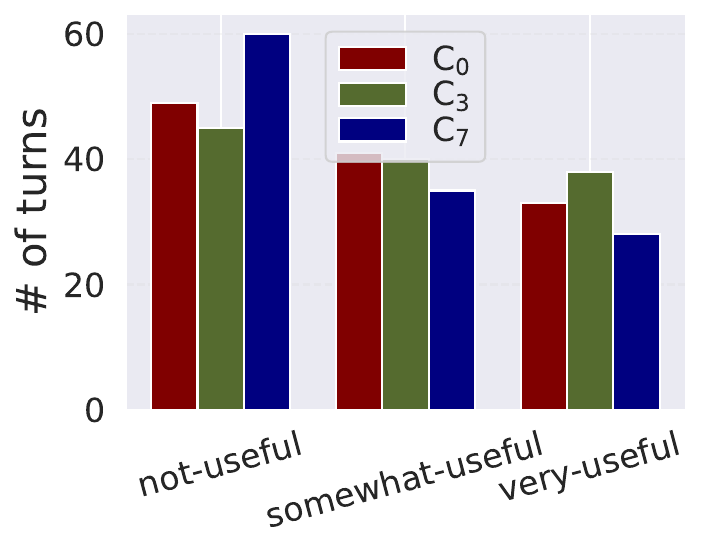}\label{fig:usefulness-phase1}}
   \caption{Distribution of (a) relevance and (b) usefulness labels for dialogue annotations in Phase~1.
   }
    \label{fig:phase1-reluse}
\end{figure}

\noindent%
\textbf{Phase~1.} Figure~\ref{fig:phase1-reluse} presents the distributions of relevance and usefulness ratings across the three variations, \Czero, \Cthree, and \Cseven. 
Figure~\ref{fig:relevant-phase1} indicates a larger number of dialogues rated as relevant when annotators had no prior context (\Czero), compared to instances of \Cthree and \Cseven, where a lower number of dialogues received such ratings. 
This suggests that in the absence of prior context, annotators are more inclined to perceive the system's response as relevant, as they lack evidence to assert otherwise. 
This trend is particularly prevalent when user utterances lean towards casual conversations, such as inquiring about a previously mentioned movie or requesting a similar recommendation to their initial query, aspects to which the annotators have no access. 
Consequently, this suggests that annotators rely on assumptions regarding the user's previous inquiries, leading to higher ratings for system response relevance.

We observe a similar trend for usefulness (Figure~\ref{fig:usefulness-phase1}), compared to \Cthree and \Cseven, \Czero has more dialogues rated as useful. The introduction of the user's next utterance introduced some level of ambiguity to annotators.
Evident in instances where the user introduced a new item not mentioned in the system's response and expressed an intention to watch it, the usefulness of the system's response became uncertain. 
This ambiguity arises particularly when annotators lack access to prior context, making it challenging to tell if the movie was mentioned before in the preceding context.

These observations highlight the impact of the amount of dialogue context on the annotators' perceptions of relevance and usefulness in Phase~1. 
This emphasizes the significance of taking contextual factors into account when evaluating \acp{TDS}.

\begin{figure}[!t]
    \centering
    
    \subfloat[Relevance]{\includegraphics[width=0.5\columnwidth]{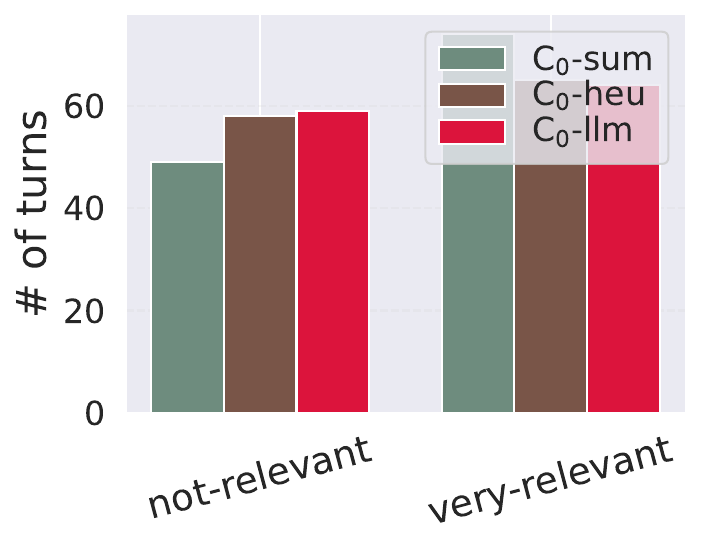}\label{relphase2}}
    \subfloat[Usefulness]{\includegraphics[width=0.5\linewidth]{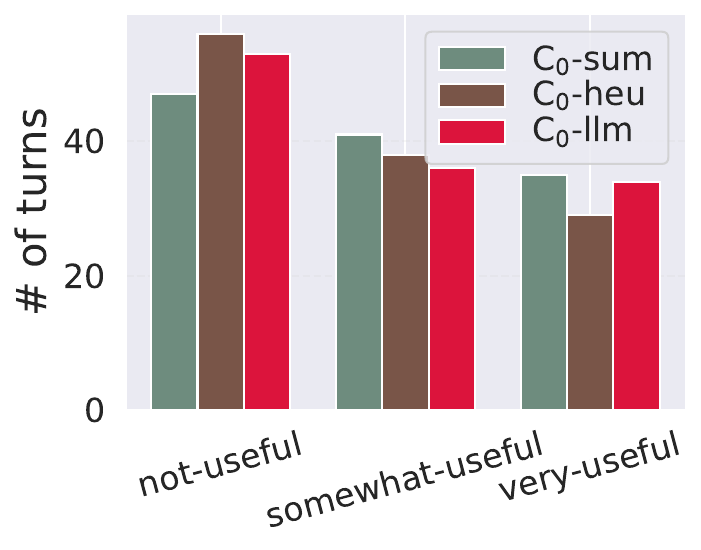}\label{usephase2}}
    \caption{Distribution of (a) relevance and (b) usefulness ratings when annotators have access to additional context in \Czero Phase~2.}
    \label{fig:distributions_phase2}
\end{figure}

\header{Phase~2} In Phase~2, we present findings on how different types of dialogue contexts influence the annotation of relevance and usefulness labels. 
When the dialogue summary is included as supplementary information for the turn under evaluation (\Czero -sum), a higher proportion of dialogues are annotated as relevant compared to \Czero -llm for relevance (60\% vs.~52.5\%, respectively); see Figure~\ref{relphase2}. 

In contrast to the observations made for relevance, we see in Figure~\ref{usephase2} that a higher percentage of dialogues are predominantly labeled as not useful when additional information is provided to the annotators. 
This accounts for 60\% in \Czero -heu, 47.5\% in \Czero -llm, and 45\% in \Czero -sum. 
This trend is consistent with our observations from Phase~1, highlighting that while system responses may be relevant, they do not always align with the user's actual information need. 
We find that \Czero -sum exhibits the highest number of dialogues rated as useful, indicating its effectiveness in providing pertinent information to aid annotators in making informed judgments regarding usefulness.

\subsection{RQ1: Effect of varying amount of dialogue context} 
\label{RQ1}

\noindent%
\textbf{Label quality.} To gauge the quality of the crowdsourced labels, we rely on inter-annotator agreement~\citep{DBLP:journals/coling/CarlettaIAA,DBLP:conf/medinfo/BoguslavIAA}.
In order to understand how the amount of dialogue context influences the quality of ratings by annotators, we calculate the agreement between annotators for both relevance and usefulness across the three variations; see Table~\ref{tab:IAA_phase1}. 
To address potential randomness in relevance ratings, given the binary scale, we randomly drop one rating from each dialogue and compute the agreement. 
We repeat this process for each annotator and calculate an average Cohen's Kappa score.
For usefulness,
\begin{table}[h]
    \caption{Inter annotator agreement~(Cohen's Kappa) and Tau correlation for relevance and usefulness across the three experimental setups in Phase~1.}
    \label{tab:IAA_phase1} 
    \centering
    \begin{tabular}{l c c c}
    \toprule
        Aspect & Variation & Kappa & Tau \\
        \midrule      
       \multirow{3}{*}{Relevance} & \Czero &  0.53 & 0.47 \\
        & \Cthree & 0.61 & 0.49 \\
        & \Cseven & 0.70 & 0.61\\
        \midrule
       \multirow{3}{*}{Usefulness} & \Czero &  0.64 & 0.54   \\
       & \Cthree & 0.68 & 0.60\\
       & \Cseven  &  0.56 & 0.41 \\
        \bottomrule
    \end{tabular}
\end{table}
we compute Kappa for each pair of annotators and then calculate the average. 
We assess the significance of the agreement using the Chi-squared method. All Kappa scores are statistically significant~($p \leq$ 0.05).

We observe an increase in the Kappa and Tau score as the dialogue context increases from \Czero to \Cseven. Despite the lack of context in \Czero, there is a moderate level of agreement regarding the relevance of the current turn. With the introduction of more context in \Cthree and \Cseven, comes an increase in agreement regarding the relevance of the current turn (see Table~\ref{tab:IAA_phase1}). 
Providing additional dialogue context seems to lead to higher levels of consensus among annotators. 
This is likely due to dataset characteristics: users tend to express their preferences early in the dialogue, rather than in subsequent exchanges.  Hence, in the case of \Czero, which only includes the current turn, when the user's utterance is incomplete, lacking an explicit expression of their preference, annotators rate more dialogues as relevant compared to \Cthree and \Cseven. 
Overall, we conclude that when annotators have insufficient information to come up with a judgment, they tend to judge the system positively, introducing a positivity bias~\citep{positivity-bias}.

We see in Table~\ref{tab:IAA_phase1}~(row 3) that despite the lack of context in \Czero, there is substantial agreement regarding the usefulness of the current turn. This is due to the availability of the user's next utterance, which serves as direct feedback on the system's response, resulting in higher agreement than for relevance assessment.
As more context is provided, there is an even higher level of agreement among annotators regarding the usefulness of the current turn. 
Access to a short conversation history significantly improves agreement on usefulness.

Surprisingly, despite having access to the entire conversation history in \Cseven, there is a slightly lower level of agreement than in \Cthree. 
The complete dialogue context may introduce additional complexity or ambiguity in determining the usefulness of the current turn. 
This occurs when conflicting feedback arises from the user's next utterance compared to the previous dialogue context. 
For example, when the system repeats a recommendation that the user has already watched or stated before, and the user expresses their intent to watch the movie in the next utterance, it leads to divergent labels. Similar trend is observed with the Tau correlations though the values are lower compared to the Kappa scores.

\begin{figure}[t]
    \centering
    
    \subfloat[Relevance]{\includegraphics[width=0.50\linewidth]{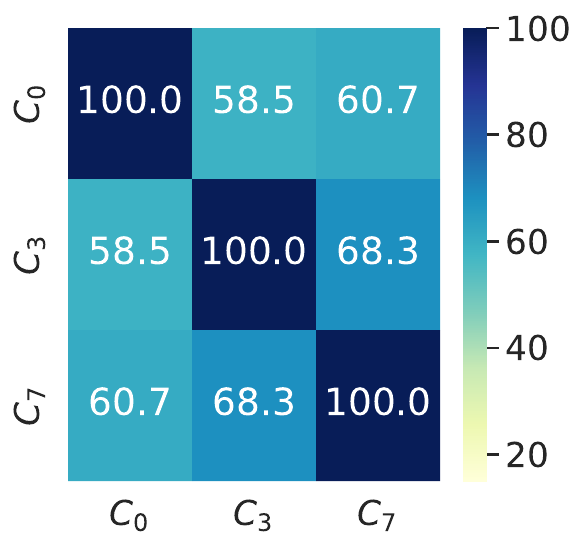}\label{relevancep1heatmap}}
    \subfloat[Usefulness]{\includegraphics[width=0.50\linewidth]{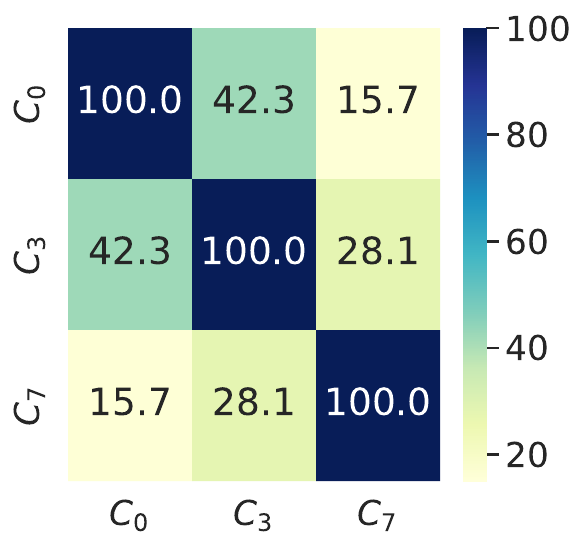}\label{usefulnessp1heatmap}}
    \caption{The percentage of agreement in (a) relevance and (b) usefulness labels across the three experimental setups in Phase~1. 
    }
    \label{fig:P1-agreementheatmap} 
\end{figure}

\header{Label consistency across conditions} We examine the impact of varying amounts of dialogue context on the consistency of crowdsourced labels across the three variations for relevance and usefulness and report the percentage of agreement in Figure~\ref{fig:P1-agreementheatmap}.
We observe moderate agreement (58.54$\%$) between annotations of \Czero and \Cthree, suggesting that annotators demonstrate a degree of consistency in their assessments when provided with different amounts of context. 
This trend continues with \Czero and \Cseven, where the agreement increases slightly to 60.98$\%$. The most notable increase is between \Cthree and \Cseven (68.29$\%$).
As annotators were exposed to progressively broader contextual information, their assessments became more consistent. 

Usefulness behaves differently. 
We observe moderate agreement (41.71\%) between \Czero and \Cthree, indicating a degree of consistency in annotator assessments within this range of context. 
A notable decrease in agreement is evident when comparing \Cthree and \Cseven, down to 28.3\%  agreement. 
The most substantial drop is observed between \Czero and \Cseven, yielding a mere 14.63\% agreement. These findings emphasize the significant impact of context on the consistency of usefulness annotations. 
For usefulness assessment providing annotators with a more focused context, improves their agreement.

With respect to \textbf{RQ1}, we note considerable differences in the labels assigned by annotators as we vary the amount of dialogue context.
As the context expands, annotators incorporate more information into their assessments, resulting in context-specific labels. 
Annotator judgments are shaped not only by response quality but also by the broader conversation. This highlights the complexity of the task and the need for a carefully designed annotation methodology that considers contextual variations. 
These findings emphasize the significance of dialogue context in annotator decision-making.

\subsection{RQ2: Effect of automatically generated dialogue context}
\label{RQ2}

\textbf{Label quality.} In Phase~2, our experiments aim to establish the impact of presenting annotators with different types of context during crowdsourcing. 
Different 
\begin{table}[t]
    \caption{Inter annotator agreement~(Cohen's Kappa) and Tau correlation for relevance and usefulness across the three experimental setups in Phase~2.}
    \label{tab:IAA_phase2}
    \centering
    \begin{tabular}{l l c c}
    \toprule
        Aspect & Variation & Kappa & Tau\\
        \midrule
       \multirow{3}{*}{Relevance}& \Czero -heu & 0.75 & 0.54 \\
       & \Czero -sum & 0.60 & 0.45 \\
       & \Czero -llm &  0.51 & 0.44\\

        \midrule
       \multirow{3}{*}{Usefulness} & \Czero -heu &  0.71 & 0.59 \\
       & \Czero -sum  & 0.63 & 0.49\\
       & \Czero -llm &  0.53 & 0.44\\
       
        \bottomrule
    \end{tabular}
\end{table}
from conventional dialogue context, we provide the annotators with the dialogue summary (\Czero -sum), the user's information need in the dialogue (\Czero -heu and \Czero -llm).
We also aim to uncover if we can improve the quality of the crowdsourced labels in \Czero to match those in \Cseven.
We calculate the Cohen's Kappa similar to Section~\ref{RQ1}; see Table~\ref{tab:IAA_phase2}.

The heuristic approach (\Czero -heu) yields the highest agreement~(Kappa and Tau), indicating a noteworthy degree of agreement in relevance assessments. 
The \ac{LLM}-generated context (\Czero -llm and \Czero -sum) results in a moderate to substantial level of agreement, signifying a reasonable level of agreement regarding the relevance of the system response. 
We observe similar results for usefulness. The heuristic approach (\Czero -heu) again leads with the highest level of agreement~(0.71 and 0.59), \Czero -sum follows with a kappa score of 0.63, while \Czero -llm has a kappa score of 0.53. This high level of agreement~(Kappa) for the two aspects indicates the quality of the labels; the additional context provided, generated either heuristically or with LLMs, is effective in conveying relevant information to annotators, leading to more consistent assessments.

For both relevance and usefulness, \Czero -heu consistently improves agreement among annotators, while the \ac{LLM}-generated context (\Czero -llm and \Czero -sum) has a substantially lower agreement than \Cseven. 
This difference reflects the limitations of \acp{LLM} in capturing context and generating a factual summary. 
While they generate coherent text, \acp{LLM} sometimes fail to correctly represent the sequential order of the dialogue and users' language patterns. 

\begin{figure}[t!]
    \centering
    
    \subfloat[Relevance]{\includegraphics[width=0.5\linewidth]{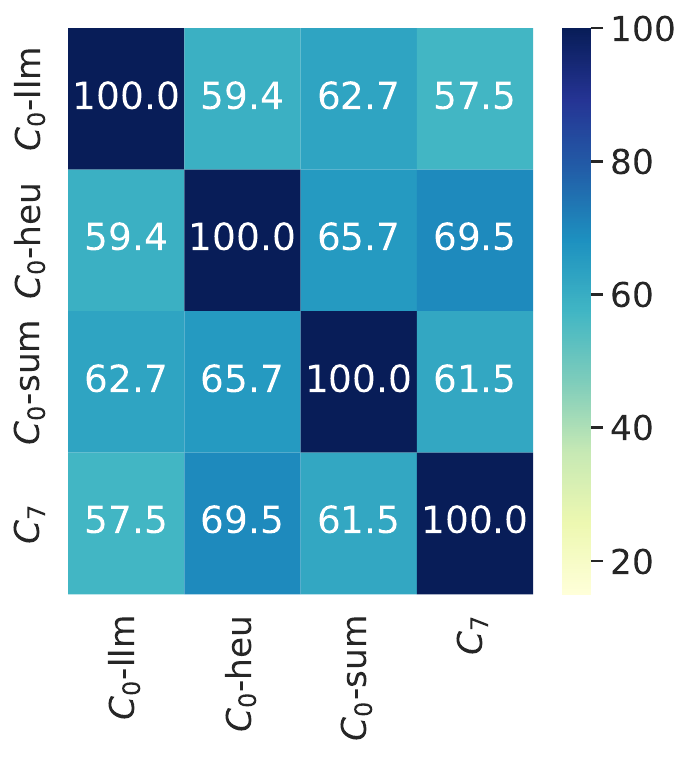}\label{fig:P2-relevanceagreementheatmap}}
    \subfloat[Usefulness]{\includegraphics[width=0.5\linewidth]{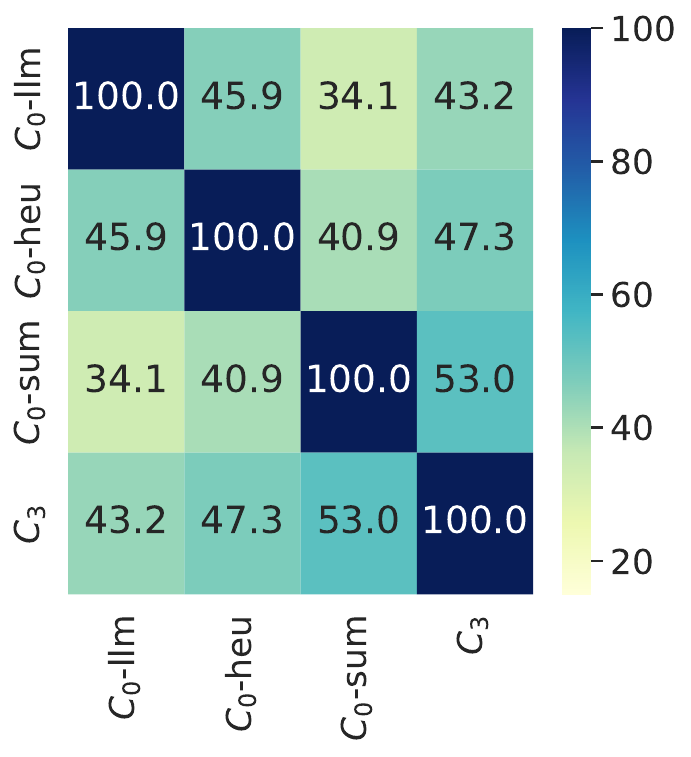}\label{fig:P2-usefulnessagreementheatmap}}
    \caption{The percentage of agreement in (a) relevance and (b) usefulness labels across the three experimental setups in Phase~2. 
    }
    \label{fig:P2-agreementheatmap}
\end{figure}

\header{Label consistency across conditions} In Figure~\ref{fig:P2-relevanceagreementheatmap} we report the agreement between the setups in Phase~2 and compare them to \Cseven~(relevance) and \Cthree~(usefulness) due to their high \ac{IAA} and label consistency.
For the relevance annotations, varying levels of agreement emerge. There is substantial agreement between \Czero -heu and \Czero -llm~(59.36\%), showing a significant overlap in the labels assigned using both methods, although there are instances where annotators differ in their assessments of relevance. 
\Czero -sum exhibits moderate label agreement with \Czero -llm~(62.74\%) and \Czero -heu~(65.67\%), pointing to relatively similar label assignments across the setups. 

We observe similar results for usefulness in Figure~\ref{fig:P2-usefulnessagreementheatmap}.
While the heuristically generated approach achieves high \ac{IAA}, the \Czero -sum method demonstrates greater consistency with all other setups in terms of usefulness.
This suggests that while annotators using the \Czero -heu approach often agreed on a single label, the chosen label may not have always been the most accurate.
We note slightly low agreement levels for a similar label between the three setups, consistent with results in Phase~1. 
Unlike relevance, which used a binary scale, usefulness was rated on a 1--3 scale.  This finer-grained scale may explain the lower agreement compared to relevance, as different types of contextual information can influence usefulness scores.

Regarding \textbf{RQ2}, we show that we can improve the consistency of the labels assigned by crowdworkers in \Czero condition by augmenting the current turn with automatically generated supplementary dialogue context. The heuristic approach demonstrates higher consistency in both \ac{IAA} and label consistency for relevance and usefulness compared to \Czero and \Cseven. 
Providing annotators with the user's initial utterance expressing their preference, particularly in scenarios lacking context, can significantly enhance the quality and consistency of crowdsourced labels. 
This approach can yield performance comparable to a setup involving the entire dialogue~\Cseven, without imposing the cognitive load of reading an entire conversation on annotators. 
This streamlines the annotation process and maintains high-quality results, offering a practical strategy for obtaining reliable labels for dialogue evaluation.

\section{Discussion and Implications}

Our findings reveal intriguing insights into the impact of context size and type on crowdsourced relevance and usefulness labels for \ac{TDS}. 
Expanding the dialogue context from \Czero to \Cseven significantly improves agreement among annotators, indicating that annotators rely on comprehensive context to make more accurate assessments. 
This trend does not hold for usefulness, where we notice a decrease in agreement when all previous dialogue context is available. 
The optimal amount of context required for reliable labels relies on the aspect evaluated.

Consistent with prior work~\citep{Eickhoff-cognitivebias,Kazai-HIT-design}, we observe an inconsistency in relevance labels across variations, with the same system response being rated differently depending on the context provided. 
Given the lack of label consistency across variations, future studies should carefully tailor their annotation task design and test various settings to ensure high-quality and consistent labels. Additionally, much care should be taken when comparing the performance of a system across several datasets when labels are crowdsourced with a different strategy to ensure a fair comparison as models similar to humans can be sensitive to the annotation strategy~\citep{annotation-sensitivity,multiannottaion}.

We also analyzed data from the open-ended question asking annotators about their experience with the annotation task.
Annotators note that dialogue summaries fail to convey a user's emotion, limiting their annotation process.
Additionally, lower accuracy of the context generated by an \ac{LLM} may lead to low agreement among annotators. 
This signifies the importance of carefully considering the quality and accuracy of generated content in the evaluation process. We provide examples in Section~\ref{supplementary-context} in the appendix.
While there may be constraints in presenting user information need and dialogue summary as dialogue context, one key consideration to take into account is the cognitive load of annotators. 
Providing a shorter, focused context reduces the cognitive burden on annotators, allowing them to devote more attention to actually evaluating a response. 
This not only streamlines the annotation process but also helps maintain high-quality results. 
Reducing the amount of content to be assessed may lead to faster annotation times without compromising the quality of ratings~\citep{DBLP:cognitiveload-ds}. Another approach to using \ac{LLM}s in annotation, is for researchers to consider co-annotation~\citep{co-annotating} between humans and \ac{LLM}s.

Optimal context varies by the aspect under evaluation, challenging the idea of a universal strategy. The consistent reliability of automatic methods suggests their potential as dependable tools for evaluation. This implies their use in generating supplementary context, eliminating the need for manual determination of context amounts. This streamlines evaluation, enhancing efficiency in context-driven evaluations for \ac{TDS}. For data lacking topic or preference shifts, heuristics perform effectively. However, \ac{LLM}s are recommended for shifting conditions, showcasing adaptability not easily discernible with heuristics.

While our primary focus was limited to relevance and usefulness, the proposed experimental design can be extended to other aspects of \acp{TDS} evaluation. 
Moreover, our findings may be task- or dataset-specific, prompting the need for further investigation into their generalizability.
As to future work, we aspire to enhance the robustness of our findings by conducting studies on larger-scale datasets. In addition following previous work by \citet{Kazai-worker-demographics,DBLP:journals/ir/KazaiKM13}, we would also want to understand the effect of annotator background: experience of interacting with conversational system or prior experience in doing the annotation task on label consistency for \acp{TDS}.

\section{Related Work}

We review related work not covered in the paper so far. 
Several user-centric dialogue evaluation metrics~\citep{mehri-eskenazi-2020-usr,Ruber-metric,Grade-metric} have been proposed. 
For \acp{TDS}, high-level dimensions such as user satisfaction~\citep{julia-understandingsat,maskari-usat} and fine-grained metrics such as relevance and interestingness~\citep{clemencia-sat} have gained interest. 
Due to the ineffectiveness of standard evaluation metrics such as ROUGE~\citep{lin-2004-rouge}, BLEU~\citep{BLEU-metric}, which show poor correlation with human judgments~\citep{Deriu2020SurveyOE}, a significant amount of research on these metrics relies on crowdsourcing dialogue evaluation labels to improve correlation with actual user ratings. 
Crowdsourcing ground-truth labels has gained momentum in \ac{IR} for tasks like search relevance evaluation~\citep{DBLP:journals/sigir/AlonsoRS08} and measuring user satisfaction in \ac{TDS}. 
A major challenge is ensuring quality and consistency of crowdsourced labels. 
Task design and annotators' behavioral features and demographics can affect the quality of the collected labels~\citep{DBLP:journals/pacmhci/Pei-behavioral,Kazai-worker-demographics,worker-bias}.
\citet{DBLP:journals/ir/KazaiKM13} examine how effort and incentive influence the quality of labels provided by assessors when making relevance judgments.
Other factors such as judgment scale~\citep{DBLP:journals/ipm/RoiteroMMS21,novikova-etal-2018-rankme}, annotator background~\citep{DBLP:conf/sigir/RoiteroSFSMD20,Kazai-worker-types}, and annotators' demographics~\citep{DBLP:conf/wsdm/DifallahFI18} have also been studied. 
Most studies focus on search systems, not dialogue systems. 
Closer to our work, \citet{DBLP:cognitiveload-ds} study the effect of cognitive bias in the evaluation of dialogue systems. 
Providing an anchor to annotators introduces anchoring bias, where annotators' ratings are close to the anchor's numerical value. 
Like~\citet{DBLP:cognitiveload-ds}, we focus on the effect of task design on the evaluation of \acp{TDS}. 
In particular, we investigate how the amount and type of dialogue context provided to annotators affect the quality and consistency of evaluation labels and the annotator experience during the evaluation task.%

\section{Conclusion}

In this work, we investigated the impact of varying the dialogue context size and type on crowdsourced evaluation labels. In particular we crowdsourced evaluation labels for two aspects: \textit{relevance} and \textit{usefulness}. Our findings reveal that optimal context is dependent on the aspect under evaluation. For relevance
annotators tend to agree more on a label when they have access to the whole dialogue context. However this does not hold for the usefulness aspect where we witness high annotator agreement when partial context is available. We show that a simple approach like providing an automatically generated user need through heuristics without revealing the entire dialogue can consistently increase annotator agreement across the two aspects. This implies that we can rely on automatic methods such as the use of LLMs to improve the productivity of the crowdworkers by reducing the amount of dialogue they have to read before evaluating the current response. 

This study contributes towards how LLMs can be integrated in the annotation process to ensure quality labels from the crowdworkers. In this work we used GPT-4 API which is not open source. For future work we will explore the use of open-source LLMs, like Llama-chat~\citep{llama2}, to facilitate a more transparent and reproducible experimental framework.

\section*{Limitations}

In this work, we dived into the effect of task design on crowdsourced evaluation labels, specifically the amount and type of context available. Nonetheless our study faces some limitations: the absence of actual user ratings hinders us from claiming an optimal strategy for presenting previous dialogue history. Despite this limitation, we highlight the noteworthy observation of high label consistency in \Cseven for relevance and \Cthree for usefulness aspect, which served as our basis for comparison. It is crucial to note that our study is exploratory in nature and thus may be data or task specific. To ensure the applicability and generalizability of our findings, it is imperative to undertake further investigations to ascertain the extent to which these findings can be extrapolated across different tasks and datasets.

\section*{Ethical Considerations}

\subsection*{Anotator diversity}
All participants in this research were master workers recruited exclusively from the United States through Amazon Mechanical Turk (MTurk). While this selection ensured a level of language proficiency and familiarity with the context, it is crucial to note that the findings of this study may not generalize universally due to the specific demographic representation. The restriction to U.S.-based annotators may introduce a limitation in terms of cultural diversity and global perspectives, influencing the external validity of the study.

\subsection*{Annotator bias}
Despite the provision of detailed instructions and examples to annotators, potential biases may still arise during the evaluation process due to the diverse backgrounds of the annotators. Cultural biases may be more pronounced if annotators from different cultural backgrounds interpret movie preferences, relevance, or usefulness in divergent ways. 
Subjective biases may also be influenced by the diverse interpretations of guidelines, as individuals from different backgrounds may have distinct views on dimensions like ``relevance'' or ``usefulness.''

To mitigate these potential biases, continuous monitoring and feedback mechanisms were incorporated into the study design.
Additionally, the study refrained from disclosing the specific research angle to annotators to prevent potential biases related to the research objectives.

\section{Acknowledgements}
This research was supported by the Dreams Lab, a collaboration between Huawei Finland, the University of Amsterdam, and the Vrije Universiteit Amsterdam, by the Hybrid Intelligence Center, a 10-year program funded by the Dutch Ministry of Education, Culture and Science through the Netherlands Organisation for Scientific Research, https://hybrid-intelligence-centre.nl, by project LESSEN with project number NWA.1389.20.183 of the research program NWA ORC 2020/21, which is (partly) financed by the Dutch Research Council (NWO), and by the FINDHR (Fairness and Intersectional Non-Discrimination in Human Recommendation) project that received funding from the European Union’s Horizon Europe research and innovation program under grant agreement No 101070212.

All content represents the opinion of the authors, which is not necessarily shared or endorsed by their respective employers and/or sponsors.

\bibliography{anthology,references}
\bibliographystyle{acl_natbib}

\clearpage
\appendix
\appendix
\section{Appendix}
In this section we provide supplementary materials used to support our main paper. These materials include: experimental conditions elaborated in Section~\ref{experimentalconditions}, quality control measures undertaken to ensure high quality crowdsourced labels and generated supplementary context in Section~\ref{datacontrol} and the prompts used to generate the supplementary context in Section~\ref{prompts}. In Section~\ref{annotation-instructions} we include the annotation instructions and screen dumps of our annotation task. Section~\ref{supplementary-context} shows sample supplementary context generated by GPT-4.

\subsection{Experimental conditions}
\label{experimentalconditions}
We list the experimental conditions used for our crowdsource experiments in Table~\ref{tab:variations}.

\begin{table*}[h]
    \caption{Descriptions of the experimental setups used for the crowdsourcing experiments with corresponding relevance and usefulness labels. Unlike relevance, usefulness includes the user's next utterance as feedback. A ``turn'' denotes a user-system exchange.}
    \label{tab:variations}
    \centering
    \begin{tabular}{l  p{13cm} }
    \toprule
        Variations  & Description \\
        \midrule
        \Czero   & Current turn with no previous dialogue context  \\
        \midrule
        \Cthree  &  Current turn with three system-user utterances as previous context  \\
        \midrule
        \Cseven &  Current turn with 7 user-system utterances as previous context \\
        \midrule
        \Czero -llm    & Current turn with an \ac{LLM}-generated user information need as dialogue context \\
        \midrule     
        \Czero -heu &  Current turn with a heuristically generated user information need as dialogue context\\
        \midrule
        \Czero -sum & Current turn with a dialogue summary as dialogue context  \\
        \bottomrule
    \end{tabular}
\end{table*}

\subsection{Data quality control} \label{datacontrol}
\header{Generated user information need and summary} To address the potential hallucination of \acp{LLM}~\citep{LLM-survey}, we implemented a quality control process for the generated user information needs and summaries, ensuring their coherence and factual accuracy.
We automatically cross-reference the movies mentioned in both the input dialogues and the summaries. A summary must contain at least two-thirds of the movies mentioned in the input dialogue to be considered valid. If this criterion is not met, the summary is discarded, and a new one is generated following the specified prompt requirements. In total, we discarded and regenerated 15 dialogue summaries.
To further ensure coherence, we randomly sampled 30$\%$ of the generated summaries and information needs. The authors reviewed them to confirm their coherence and alignment with the information presented in the input dialogue. This process enhanced the quality and reliability of the generated content.

\header{Crowdsourced labels} To ensure a high quality of the collected data, we incorporated attention-checking questions into the \ac{HIT}. Annotators were required to specify the number of utterances in the dialogues they were evaluating and to identify the last movie mentioned in the system response being evaluated. 10$\%$ of the \acp{HIT} were rejected and returned back to collect new labels.
In total, we gathered 1440 data samples from the crowdsourcing task, spanning six variations for relevance and usefulness. We employed majority voting to establish the final relevance and usefulness dialogue label. 

\subsection{Prompts}\label{prompts}
In Table~\ref{tab:prompts} we show the final prompts used to generate the user information and dialogue summary with GPT-4.

\begin{table*}[!h]
    \caption{Prompts used to generate the supplementary context; user information need and dialogue summary with GPT-4.}
    \label{tab:prompts}
    \centering
    \begin{tabular}{|p{0.97\linewidth}|}
        \hline
        \textbf{Dialogue summary prompt} \\
        \hline
        Below you are provided with dialogues between a user and the system about movie recommendations. Generate a complete short and informative summary extractively which is half the length of the dialogue. \\
        \hline
        \textbf{User information need prompt} \\
        \hline
        Given the following user and system dialogue in a movie recommendation conversation, generate a concise user's goal in a natural manner. State only the goal without extra text. Start the sentence with ``the user wants.'' \\
        \hline
    \end{tabular}
\end{table*}

\subsection{Annotation instructions and screen dumps}\label{annotation-instructions}

\begin{figure*}[!ht]
    \centering
    \includegraphics[width=1\linewidth]{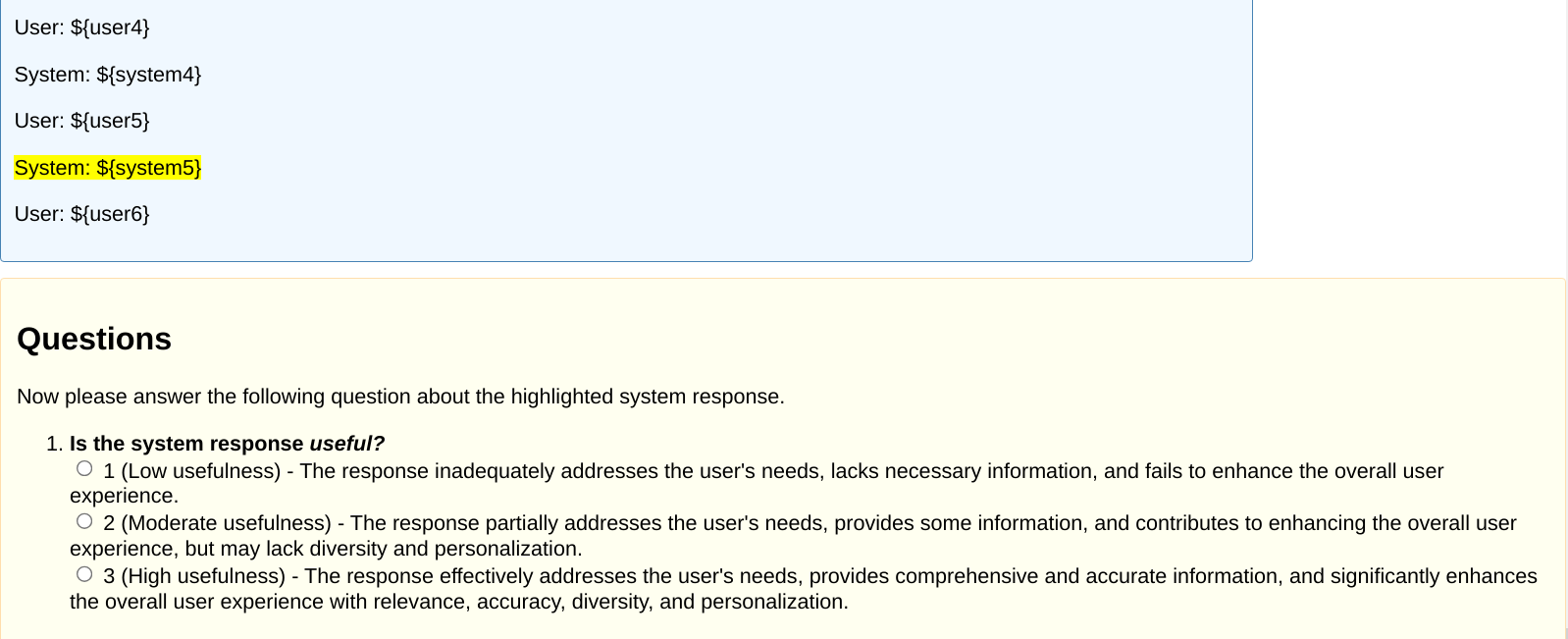}
    \caption{Annotation interface for phase 1 when evaluating response usefulness for \Cthree}
    \label{fig:phase1-ann}
\end{figure*}

\begin{figure*}[!h]
    \centering
    \includegraphics[width=1\linewidth]{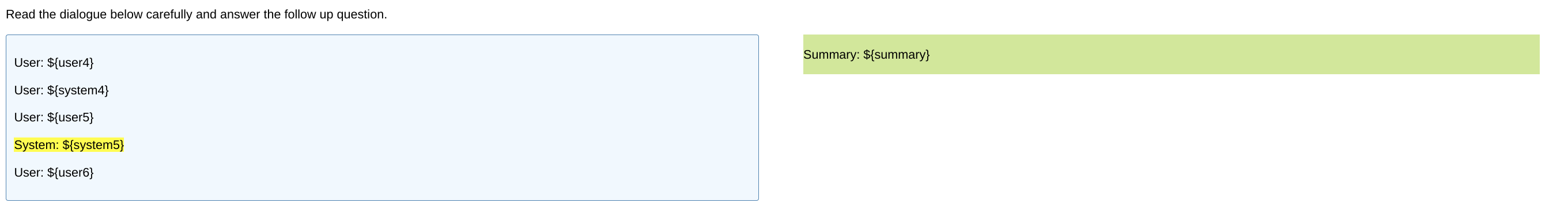}
    \caption{Annotation interface for phase 2 when evaluating response usefulness with supplementary context}
    \label{fig:phase2-ann}
\end{figure*}

Table~\ref{tab:survey-intro} details the annotation instructions for the relevance and usefulness evaluations. In Figure~\ref{fig:phase1-ann} and \ref{fig:phase2-ann} we show the annotation interface used for Phase~1 and Phase~2, respectively.

\begin{table*}
    \caption{Annotation instructions provided to the annotators for relevance evaluation. The instructions are the same for usefulness apart from the aspect being evaluated.}
    \label{tab:survey-intro}
    \centering
    \begin{tabular}{|p{0.97\linewidth}|}
        \hline
        \textbf{Introduction} \\
        \hline
        Thank you for helping us out! Below we explain everything in full detail. Please make sure to read the instructions carefully. \\
        \hline
        \textbf{Purpose} \\
        \hline
        The aim of this survey is to evaluate the quality of a system's response. We want to evaluate the dialogue system's performance and gather insights for improvements. We will ask you to evaluate the system response on one metric, that we will discuss in more detail below. \\
        \hline
        \textbf{Scenario Outline} \\
        \hline
        Imagine you are evaluating a dialogue system that generates a response to user queries. Your task is to assess the response based on relevance. We will provide examples and detailed explanations of this criteria below. \\
        \hline
        \textbf{Task} \\
        \hline
        In each HIT, you will be presented with a dialogue chunk. Your task is to evaluate the last system response based on the given criteria. Please review the explanations and examples for the criteria to ensure your understanding before proceeding with the evaluation. Keeping the scenario that was outlined above in mind, we would like to ask you to judge the system response on relevance. \\
        \hline
    \end{tabular}
\end{table*}

\subsection{Sample supplementary context}\label{supplementary-context}
In Table~\ref{tab:movie-dialogues-summary} we show sample user information need and summary generated by GPT-4.

\begin{table*}[h]
\caption{Sample dialogue summaries as supplementary context generated by GPT-4.}
    \centering
    \begin{tabular}{|p{0.97\linewidth}|}
        \hline
        \textbf{Dialogue 1} \\
        \hline
        User inquires about a good family movie recommendation similar to "Real Steel (2011)" or "The Lego Movie (2014)". System recommends "Super (2010)", an action-comedy about a regular guy who becomes a self-made superhero, describing it as hilarious and entertaining. The user shows interest in this recommendation. \\
        \hline
        \textbf{Dialogue 2} \\
        \hline
        The user asked for coming-of-age movie recommendations and mentioned they enjoyed "My Girl (1991)" and "Lucas (1986)". The system suggested watching "The Spectacular Now (2013)", a film where Shailene Woodley stars as a character who forms a bond with a troubled classmate. \\
        \hline
        \textbf{Dialogue 3} \\
        \hline
        User seeks a dramatic love story to watch. System recommends "The Notebook (2004)", but the user has watched it, as well as "Titanic (1997)". Both films are favored by the user; they desire to watch something new. \\
        \hline
        \textbf{Dialogue 4} \\
        \hline
        The user requests animated movie recommendations following their enjoyment of "The Incredibles (2004)". The system suggests other movies, including "Monsters, Inc. (2001)" and its sequel "Monsters University (2013)", which the user approves. The conversation pivots to the topic of successful sequels, citing "Toy Story 3 (2010)" as an example despite the user's disagreement, favoring the original movie, "Toy Story (1995)". \\
        \hline
        \textbf{Dialogue 5} \\
        \hline
        The user wants to find a thrilling crime movie like "Thor: Ragnarok (2017)" for their weekend. The system suggested they watch "The Snowman (2017)" but the user declined. However, the system then gave another recommendation, "First Kill (2001)". \\
        \hline
    \end{tabular}
    
    \label{tab:movie-dialogues-summary}
\end{table*}

\end{document}